\documentclass{article}
\usepackage{spconf,amsmath,graphicx,url}
\usepackage{multirow}
\usepackage{booktabs}
\usepackage{algorithm}
\usepackage{algorithmicx}
\usepackage{algpseudocode} 
\usepackage{amsmath}
\usepackage{amssymb}
\usepackage{bm}

\title{TARGETDROP: A TARGETED REGULARIZATION METHOD FOR \\CONVOLUTIONAL NEURAL NETWORKS}
\name{Hui Zhu\textsuperscript{\textit{1,2}}, Xiaofang Zhao\textsuperscript{\textit{1,}}\sthanks{Corresponding author of this work.}}

\address{\textsuperscript{1}Institute of Computing Technology, Chinese Academy of Sciences, China\\\textsuperscript{2}University of Chinese Academy of Sciences, China}
\begin{document}
%
\maketitle
\begin{abstract}
Dropout regularization has been widely used in deep learning but performs less effective for convolutional neural networks since the spatially correlated features allow dropped information to still flow through the networks. Some structured forms of dropout have been proposed to address this but prone to result in over or under regularization as features are dropped randomly. In this paper, we propose a targeted regularization method named TargetDrop which incorporates the attention mechanism to drop the discriminative feature units. Specifically, it masks out the target regions of the feature maps corresponding to the target channels. Experimental results compared with the other methods or applied for different networks demonstrate the regularization effect of our method.

\end{abstract}
\section{Introduction}
Convolutional neural networks are widely used in the field of computer vision and have achieved great success. Many excellent neural architectures have been designed successively such as ResNet \cite{resnet}, DenseNet \cite{densenet} and SENet \cite{senet}. In order to solve the over-fitting problem caused by the increase in the number of parameters for convolutional neural networks, many regularization methods have been proposed, such as weight decay, data augmentation and dropout \cite{dropout}. 

However, The effect of dropout for convolutional neural networks is not as significant as that for fully connected networks because the spatially correlated features allow dropped information to still flow through convolutional networks \cite{dropblock}. To address this problem, some structured forms of dropout have been proposed such as SpatialDropout \cite{spatialdropout}, Cutout \cite{cutout} and DropBlock \cite{dropblock}. But these methods prone to result in over or under regularization as features are dropped randomly. 

Some methods attempt to combine structured dropout with attention mechanism such as AttentionDrop \cite{attentiondrop} and CorrDrop \cite{corrdrop}. However, these methods only mask out the units with higher activation values or the regions with less discriminative information in the spatial dimension. They ignore the instructive information in the channel dimension which is proven to be meaningful in convolutional neural networks \cite{senet}, even in dropout-based regularization methods\cite{dropblock}.

In this paper, we propose a novel regularization method named TargetDrop, which drops the feature units with a clear target. Specifically, we choose the target channels and then drop the target regions in the corresponding feature maps. As is shown in Fig.1, compared with naive Dropout and DropBlock which may lead to unexpected results by dropping randomly, our TargetDrop prone to precisely mask out several effective features of the main object, thus forcing the network to learn more crucial information. Our experimental results demonstrate that TargetDrop can greatly improve the performance of convolutional neural networks and outperforms many other state-of-the-art methods on public datasets CIFAR-10 and CIFAR-100 which we attribute to our method. 

Our contributions are summarized as follows:

\noindent $\bullet$ We propose a targeted regularization method which incorporates attention mechanism to address the problem for unexpected results caused by dropping randomly.\\
\noindent $\bullet$ We propose the rule of choosing target channels and target regions, and further analyse the regularization effect. \\
\noindent $\bullet$ Our method achieves better regularization effect compared with the other state-of-the-art methods and is applicable to different architectures for image classification tasks.

\begin{figure}
  \centering
  \includegraphics[width=1\linewidth]{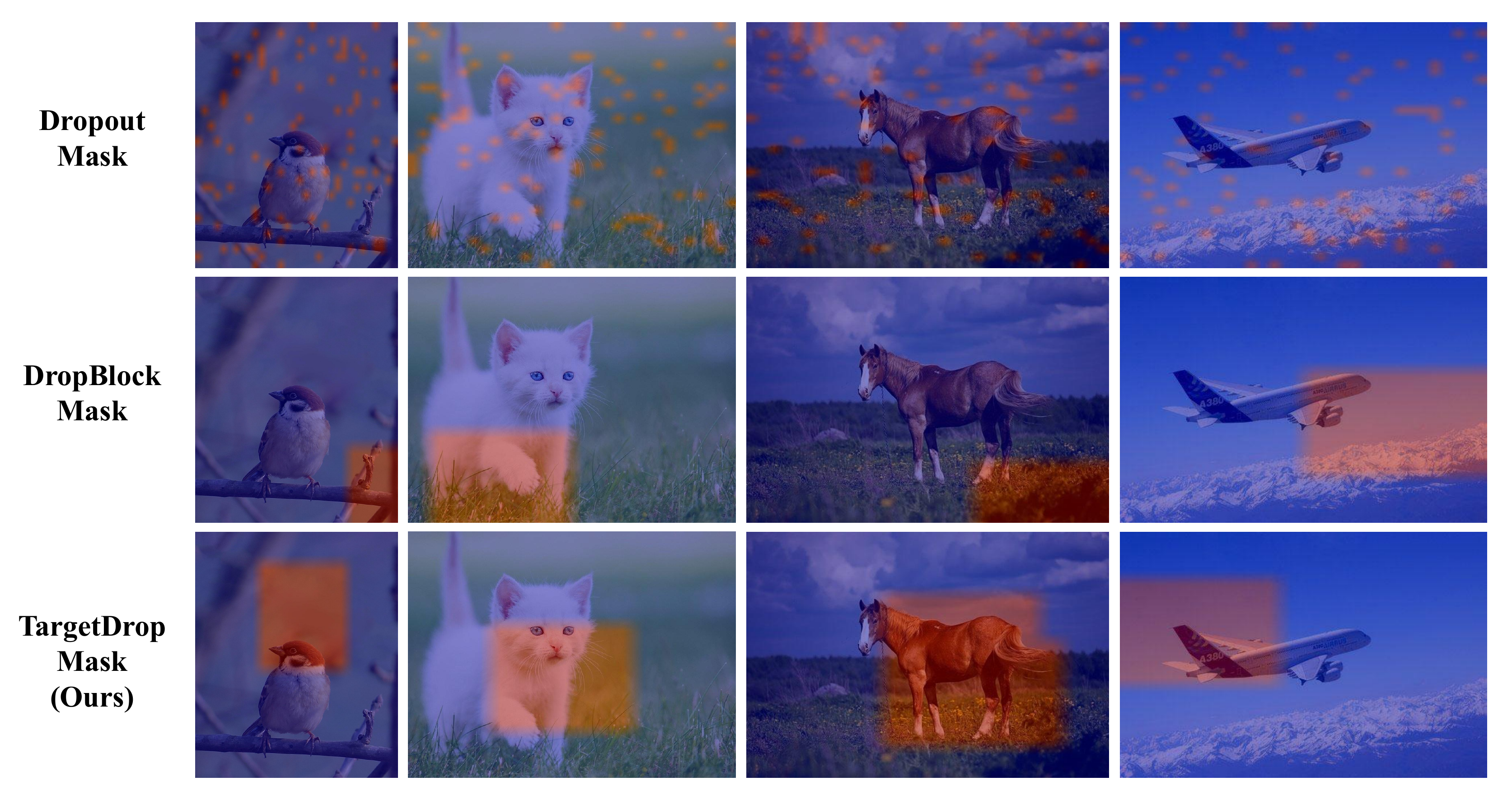}
  \caption{Masks of naive Dropout \cite{dropout}, Dropblock \cite{dropblock} and our TargetDrop. The red regions denote the regions to be masked.}
\end{figure}

\begin{figure*}
  \centering
  \includegraphics[width=1\linewidth]{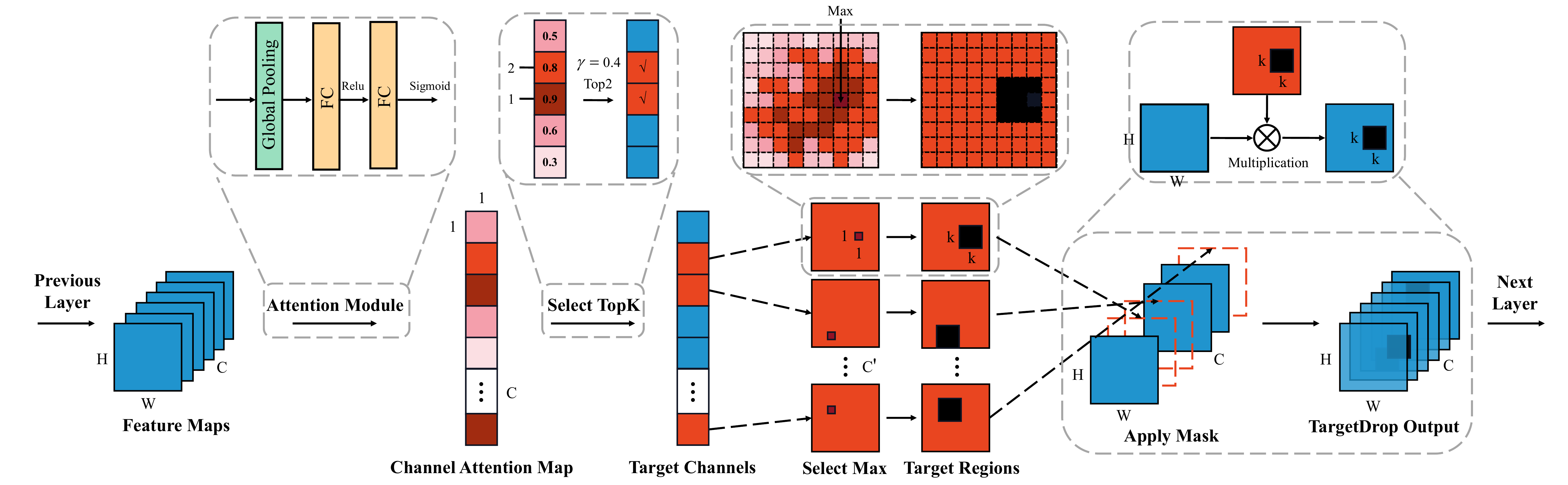}
  \caption{The pipeline of TargetDrop. First, channel attention map is generated by processing the output of the previous layer through attention mechanism. Next, $topK$ elements are selected as the target channels according to the drop probability $\gamma$. Then, locate to a pixel with the maximum value in the feature map corresponding to each target channel and generate a mask by dropping the $k \times k$ target region. Finally, the mask is applied to the original feature maps by the multiplication operation.}
\end{figure*}

\section{Related Work}
Since Dropout \cite{dropout} was proposed to improve the performace of networks by avoiding overfitting the training data, a series of regularization variations have been proposed such as DropConnect \cite{dropconnect}, SpatialDropout \cite{spatialdropout}, DropPath \cite{droppath}, DropBlock \cite{dropblock}, AttentionDrop \cite{attentiondrop}, CorrDrop\cite{corrdrop} and DropFilterR \cite{dropfilterR}. In addition, several methods about attention processing are also related as our method incorporates the attention mechanism into the dropout-based regularization. Methods for computing the spatial or channel-wise attention have achieved a certain effect such as Squeeze-and-Excitation (SE) module \cite{senet}, Convolutional Block Attention Module (CBAM) \cite{cbam} and Selective Kernel (SK) unit \cite{sknet}. Our method outperforms the dropout-based regularization counterparts by utilizing the attention mechanism to achieve the targeted dropout.

\section{Methods}
In this section, we propose our method TargetDrop which mainly contains seeking out the target channels and target regions. The pipeline of TargetDrop is shown in Fig. 2.

\subsection{Target Channels}
Given the output of the previous convolutional layer as $\bm{U} = \left[ \bm{u_{1}}, \bm{u_{2}},\cdots,\bm{u_{C}} \right] \in \mathbb{R}^{H \times W \times C}$, where $H$ and $W$ are the height and width of the feature map respectively, $C$ is the number of channels. As a first step, we are eager to figure out the importance of each channel. We aggregate the spatial information of each feature map into channel-wise vector by using global average pooling which has been proven to be effective \cite{senet, cbam}. This vector $\bm{v} \in \mathbb{R}^{1 \times 1 \times C}$ can be regarded as the statistic generated by shrinking through spatial dimensions $H \times W$ and this operation $\bm{F}_{\bm{U} \rightarrow \bm{v}}$ can be defined as:

\begin{equation}
v_{c} = \bm{F}_{\bm{U} \rightarrow \bm{v}}(\bm{u_{c}}) = \frac{1}{H W} \sum_{i=1}^{H}\sum_{j=1}^{W} u_{c}(i,j)
\end{equation}
where $v_{c}$ denotes the $c$-th element of $\bm{v}$. To further capture channel-wise dependencies, the vector is then forwarded to a shared network to produce the channel attention map $\bm{M} \in \mathbb{R}^{1 \times 1 \times C}$. The shared network is composed of two fully connected (FC) layers and two activation functions. Specifically, a dimensionality-reduction layer with parameters $\bm{W}_{1} \in \mathbb{R}^{\frac{C}{r} \times C}$, a ReLU, a dimensionality-increasing layer with parameters $\bm{W}_{2} \in \mathbb{R}^{C \times \frac{C}{r}}$ and then a Sigmoid function are
connected alternately. Here, $r$ is the reduction ratio to adjust the bottleneck. The map indicates the inter-channel relationships, and this operation $\bm{F}_{\bm{v} \rightarrow \bm{M}}$ can be defined as:

\begin{equation}
\bm{M} = \bm{F}_{\bm{v} \rightarrow \bm{M}}(\bm{v},\bm{W}) = \sigma \left( \bm{W}_{2} \delta \left(\bm{W}_{1}\bm{v} \right)\right)
\end{equation}
where $\delta$ and $\sigma$ refer to the ReLU and Sigmoid, respectively. 

Then, we sort all the values in $\bm{M}$ and select the elements (tag "1" means to be selected and "0" if not) with top $K$ values as the target according to the drop probability $\gamma$. Specifically, the channels corresponding to those elements marked as tag "1" in the vector $\bm{T} \in \mathbb{R}^{1 \times 1 \times C}$ are the target channels. Given the top $K$-th value in $\bm{M}$ as $M_{topK}$ and this process can be described as:

\begin{equation}
K = \lfloor \gamma C \rfloor, \qquad \quad
T_{p} =
\left\{  
	\begin{array}{lr}
	1 \quad & M_{p} \geq M_{topK} \\ 
	0 \quad & \rm otherwise
	\end{array}
\right. 
\end{equation}
where $M_{p}$ and $T_{p}$ denote the $p$-th elements of $\bm{M}$ and $\bm{T}$. Based on this, we further select the target regions of the original $H \times W$ feature maps corresponding to the target channels which we will elaborate in the following subsection.

\subsection{Target Regions}
For each feature map corresponding to a target channel, we hope to further seek out a region with  much discriminative information in convolution operation. Utilizing spatial attention mechanism like a convolution operation with the kernel size of 7$\times $7 is not necessary and may lead to considerable additional computation. Considering the continuity of image pixel values \cite{attentiondrop}, we can simply locate to a pixel with maximum value and the other top values distributed in the surrounding continuous regions are most likely to be certain crucial features of the main object. Hence the location $(a, b)$ with the maximum value will be selected and the $k \times k$ region centered around it will be dropped. $h_{1}$, $h_{2}$, $w_{1}$ and $w_{2}$ represent the boundaries of the target region and the TargetDrop mask $\bm{S} = \left[ \bm{s_{1}}, \bm{s_{2}},\cdots,\bm{s_{C}} \right] \in \mathbb{R}^{H \times W \times C}$ can be described as: 

\begin{equation}
\begin{array}{c}
h_{1}=a - \lfloor \frac{k}{2} \rfloor, h_{2}=a + \lfloor \frac{k}{2} \rfloor \smallskip \\

 w_{1}=b - \lfloor \frac{k}{2} \rfloor, w_{2}=b + \lfloor \frac{k}{2} \rfloor
 \end{array}
\end{equation}

\begin{equation}
s_{q}(m,n) =
\left\{  
	\begin{array}{lr}
	0 & T_{q}=1 \wedge h_{1} \leq m \leq h_{2} \wedge w_{1} \leq n \leq w_{2}\\
	1 & \rm otherwise
	\end{array}
\right. 
\end{equation}
where $s_{q}$ and $T_{q}$ denote the $q$-th elements of $\bm{s}$ and $\bm{T}$. Given the final output as $\bm{\widetilde{U}}= \left[ \bm{\widetilde{u}_{1}}, \bm{\widetilde{u}_{2}},\cdots,\bm{\widetilde{u}_{C}} \right] \in \mathbb{R}^{H \times W \times C}$. Finally, we apply the mask and normalize the features:
\begin{equation}
\bm{\widetilde{u}_{z}} = \bm{u}_{z} \odot \bm{s}_{z} \times \frac{numel(\bm{s}_{z})}{sum(\bm{s}_{z})}
\end{equation}
where $\bm{u}_{z}$ and $\bm{s}_{z}$ denote the $z$-th elements of $\bm{u}$ and $\bm{s}$, $numel(\bm{s}_{z})$ counts the number of units in $\bm{s}_{z}$, $sum(\bm{s}_{z})$ counts the number of units where the value is "1" and $\odot$ represents the point-wise multiplication operation.

\subsection{TargetDrop}
\begin{algorithm}[htb]
  \caption{TargetDrop}
  \begin{algorithmic}[1]
    \Require
    Output of the previous layer: $\mathcal \bm{U}$, Drop probability: $\mathcal \gamma$, Size of the dropped block: $k \times k$, Phase of run: $phase$
    \Ensure  
   Final feature maps: $\mathcal \bm{\widetilde{U}}$.
      \If {$phase$ == $interface$}  
      	\State \Return{$\mathcal \bm{U}$}
      \EndIf 
      \State Generate the channel attention map $\mathcal M$;
      \State Select the top $\mathcal K$-th value $\mathcal M_{top \mathcal K}$ in $\mathcal M$ according to $\mathcal \gamma$ and generate the target channels: $\mathcal T$;
      \State Locate to the maximum value $u(a, b)$, produce a mask with the center being it and the width, height being $k$: $\mathcal S$;
      \State Apply the mask $\mathcal S$ by the multiplication operation and normalize the features to generate the output: $\mathcal \bm{\widetilde{U}}$;
   \State \Return{$\mathcal \bm{\widetilde{U}}$}
  \end{algorithmic}  
\end{algorithm}

\noindent The pseudocode of our method is described in Algorithm 1.

\begin{table}
  \caption{Comparison against the results of different state-of-the-art dropout-based regularization methods for classification accuracy of ResNet-18 on CIFAR-10 and CIFAR-100.}\smallskip
  \centering
  \resizebox{1\columnwidth}{!}{
  \begin{tabular}{l|c|c}
    \toprule
    \midrule
    \textbf{Methods}& \textbf{C10 Error}(\%) &\textbf{C100 Error}(\%)\\
	\midrule
	\midrule
    No Regularization & 4.72 & 22.46 \\
    \midrule
    Dropout \cite{dropout}  & 5.14 & 23.82  \\
    \midrule
    DropBlock \cite{dropblock} & 4.59 & 21.95  \\
    \midrule
    AttentionDrop \cite{attentiondrop} & 4.51 & 21.53  \\
    \midrule
    TargetDrop (Ours)  & \textbf{4.41} & \textbf{21.37}  \\
    \midrule
    \midrule
    Cutout \cite{cutout} & 3.99 & 21.96  \\ 
    \midrule
    Cutout + TargetDrop (Ours)  & \textbf{3.67} & \textbf{21.25}  \\
    \midrule
    \bottomrule
  \end{tabular}
  }
\end{table}

\begin{table}
  \caption{The regularization effect of TargetDrop on CIFAR-10 with different convolutional neural networks.}\smallskip
  \centering
  \resizebox{1\columnwidth}{!}{
  \begin{tabular}{l|c|c|c|c}
    \toprule
    \midrule
    \multirow{2}{*}{\textbf{Networks}} & \multirow{2}{*}{\textbf{Params}} & \multicolumn{3}{c}{\textbf{C10 Test Error}(\%)}\\
    \cmidrule{3-5}
    & (Mil.) & Baseline & Dropout \cite{dropout} & TargetDrop (Ours)\\
    \midrule
	\midrule
    ResNet-20 \cite{resnet}& 0.27 & 8.21 &7.80 & \textbf{7.61}\\
    \midrule
	VGG-16 \cite{vgg}  & 14.73 & 6.17 & 6.43  & \textbf{5.89} \\
    \midrule
    WRN-28-10 \cite{wrn} & 36.48 & 4.02 & 4.04 & \textbf{3.68} \\
    \midrule
    \bottomrule
  \end{tabular}
  }
\end{table}

\section{Experiments}
In this section, we introduce the implementation of experiments and report the performance of our method. We compare TargetDrop with the other state-of-the-art dropout-based methods on CIFAR-10 and CIFAR-100 \cite{cifar} and apply it for different architectures. We further analyse the selection of hyper-parameters and visualize the class activation map.

\subsection{Datasets}
We use CIFAR-10 and CIFAR-100 \cite{cifar} for image classification as basic datasets in our experiments. For preprocessing, we normalize the images using channel means and standard deviations and apply a standard data augmentation scheme: zero-padding the image with 4 pixels on each side, then cropping it to $32\times32$ and flipping it horizontally at random.

\subsection{Training Method}
Networks using the official PyTorch implementation are trained on the full training dataset until convergence and we report the highest test accuracy following common practice. The hyper-parameters in our experiments are as follows: the batch size is 128, the optimizer is SGD with Nesterov’s momentum of 0.9, the initial learning rate is 0.1 and is decayed by the factor of 2e-1 at 0.4, 0.6, 0.8 ratio of total epochs. For Cutout \cite{cutout}, which is used in a few experiments, the cutout size is 16$\times$16 for CIFAR-10 and 8$\times$8 for CIFAR-100.

\subsection{Experiments for TargetDrop and Results}
The experiments for TargetDrop we conducted mainly contain two parts: comparing the regularization effect with the other state-of-the-art dropout-based methods for ResNet-18 and show the performance for different architectures. We demonstrate the experiments and results in detail from these two aspects in the following two paragraphs. Specifically, in this part, the reduction ratio $r$ is 16, the drop probability and block size for TargetDrop are 0.15 and 5, respectively. 

\noindent \textbf{Comparison against the results of other methods.} We compare the regularization effect with the other state-of-the-art dropout-based methods on ResNet-18. We apply our TargetDrop in the same way with the other methods that adding the regulation to the outputs of first two groups for a fair comparison. Specially, several methods we reproduced can not reach the reported results, so we refer to the data in the original paper \cite{attentiondrop} directly for these. As is shown in Table 1, the results of our method outperform the other methods on CIFAR-10 and CIFAR-100. Moreover, combined with Cutout \cite{cutout}, our method can achieve better regularization effect.

\noindent \textbf{Regularization on different architectures.} We further conduct experiments on CIFAR-10 with several classical convolutional neural networks to demonstrate that our method is applicable to different architectures. As is shown in Table 2, our method TargetDrop can improve the performances of different architectures. We can notice that Dropout \cite{dropout} is not applicable for convolutional neural networks which is mentioned above and TargetDrop is an effective dropout-based regularization method for the networks on different scales.

\begin{figure}
  \centering
  \includegraphics[width=1\linewidth]{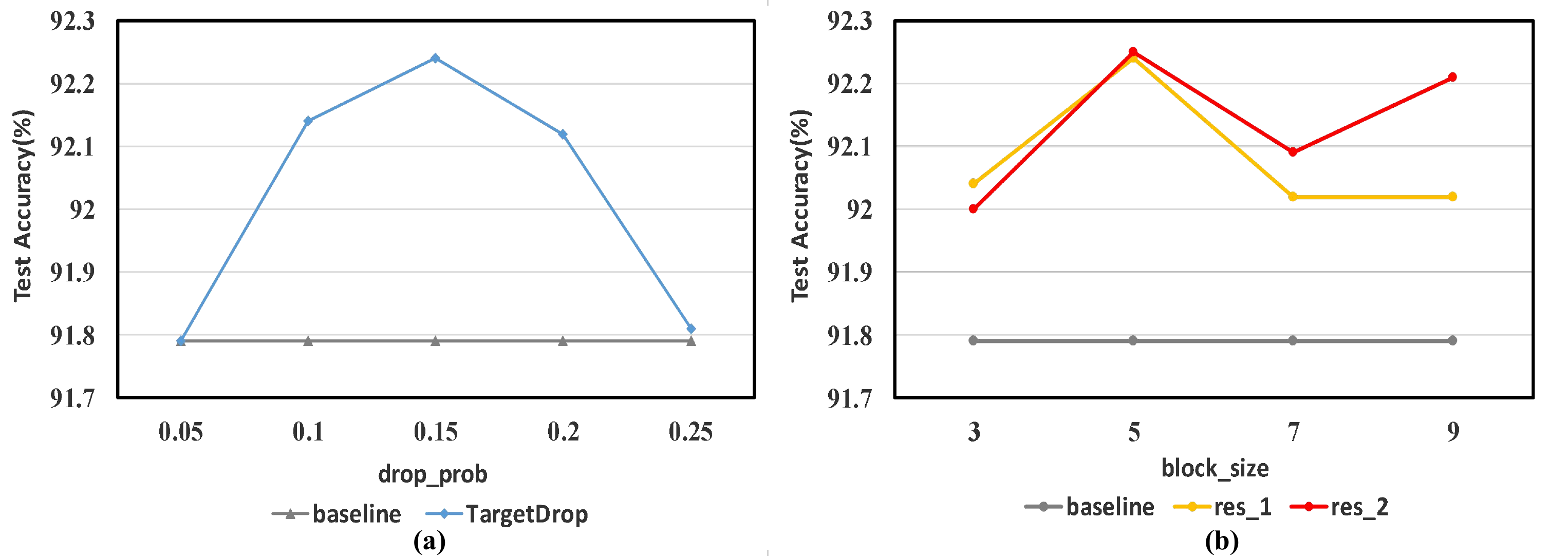}
  \caption{(a) and (b) are the test accuracy on CIFAR-10 with different drop probabilities and block sizes, respectively.}
\end{figure}

The number of parameters added in our method is presented as follows. The additional parameters come from the channel attention mechanism, and the additional computation includes the simple selection of the maximum pixel besides this. The number of parameters in the training process only increase by about 0.02\% and the amount of computation increase similarly. While in the test process, TargetDrop is closed like other methods, so the complexity will not change.

\subsection{Analysis of the Hyper-parameters Selection}
In this subsection, we further analyse the selection of hyper-parameters mentioned above: the drop probability $\gamma$ and the block size $k$. To analyse the former, we constrain the block size to 5 and TargetDrop is applied to the output of the first group (the size of the feature map is 32$\times$32). To analyse the latter, we constrain the drop probability to 0.15 and TargetDrop is applied to the outputs of the first two groups (the size of the feature map is 32$\times$32 for res\_1 and 16$\times$16 for res\_2). As is shown in Fig. 3, our method is suitable for more channels and insensitive to different hyper-parameters to a certain extent which may be due to the targeted dropout. The drop probability of 0.15 and the block size of 5 are slightly better.

\begin{figure}
  \centering
  \includegraphics[width=1\linewidth]{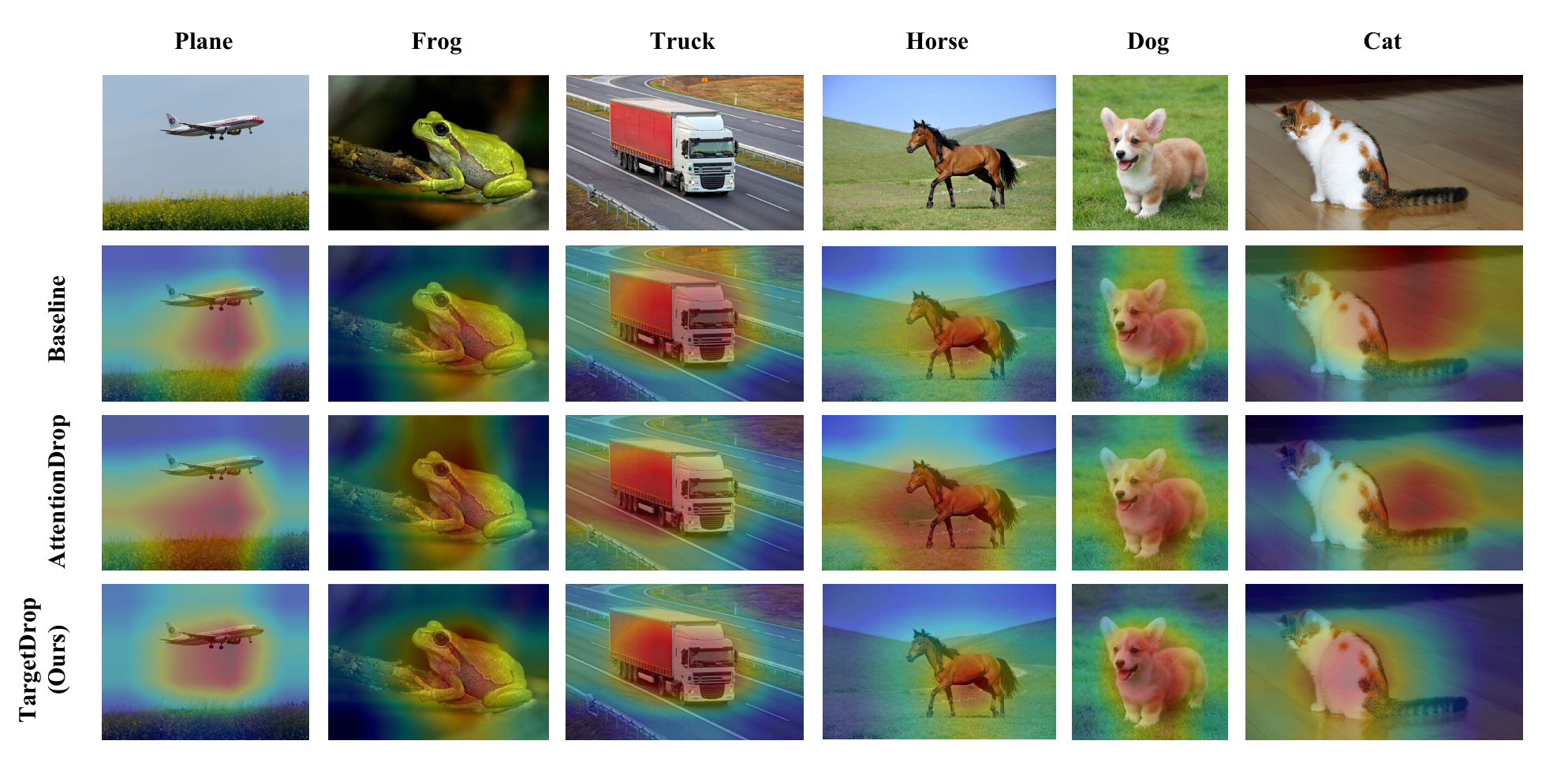}
  \caption{Class activation mapping(CAM) \cite{cam} for ResNet-18 model trained with no regularization, AttentionDrop \cite{attentiondrop} and our method TargetDrop.}
\end{figure}

\subsection{Activation Visualization}
In this subsection, we utilize the class activation mapping (CAM) \cite{cam} to visualize the activation units of ResNet-18 \cite{resnet} on several images as shown in Fig. 4. We can notice that the activation map generated by model regularized with our method TargetDrop demonstrates strong competence in capturing the extensive and relevant features towards the main object. Compared with the other methods, the model regularized with TargetDrop tends to precisely focus on those discriminative regions for image classification which we attribute to targeting and masking out certain effective features corresponding to the crucial channels.

\section{Conclusion}
In this paper, we propose the novel regularization method TargetDrop for convolutional neural networks, which addresses the problem for unexpected results caused by the untargeted methods to some extent by considering the importance of the channels and regions of feature maps. Extensive experiments demonstrate the outstanding performance of TargetDrop by comparing it with the other methods and applying it for different architectures. Furthermore, we analyse the selection of hyper-parameters and visualize the activation map to prove the rationality of our method. In addition to image classification tasks, we believe that TargetDrop is  suitable for more datasets and tasks in the field of computer vision. 

\bibliographystyle{IEEEbib}
\bibliography{strings,refs}

\end{document}